\documentclass{bmvc2k}

\title{Towards Unsupervised Sketch-based Image Retrieval}

\addauthor{Conghui Hu}{conghui@nus.edu.sg}{1}
\addauthor{Yongxin Yang}{yongxin.yang@ed.ac.uk}{2}
\addauthor{Yunpeng Li}{yunpeng.li@surrey.ac.uk}{3}
\addauthor{Timothy M. Hospedales}{ t.hospedales@ed.ac.uk}{2}
\addauthor{Yi-Zhe Song}{y.song@surrey.ac.uk}{3}

\addinstitution{
 National University of Singapore\\
 Singapore
}
\addinstitution{
 University of Edinburgh\\
 United Kingdom
}
\addinstitution{
 University of Surrey\\
 United Kingdom
}

\runninghead{Hu et al.}{Towards Unsupervised Sketch-based Image Retrieval}

\def\eg{\emph{e.g}\bmvaOneDot}

\usepackage{times}
\usepackage{epsfig}
\usepackage{graphicx}
\usepackage{amsmath}
\usepackage{amssymb}
\usepackage{makecell}
\usepackage{multirow}
\usepackage[ruled,vlined]{algorithm2e}
\usepackage{algorithmic}
\usepackage{color}
\usepackage{amssymb}
\usepackage{pifont}
\usepackage{comment}

\begin{document}

\newcommand{\superkeypoint}[1]{\vspace{0.05em}\noindent\textit{\textbf{#1}}\quad}
\newcommand{\keypoint}[1]{\vspace{0.05em}\noindent\textbf{#1}\quad}
\newcommand{\doublecheck}[1]{\textcolor{black}{#1}}
\newcommand{\conghuiadded}[1]{\textcolor{black}{#1}}
\newcommand{\chaddednew}[1]{\textcolor{blue}{#1}}

\newcommand{\modelName}[1]{UCRAL}
\newcommand{\cut}[1]{}
\renewcommand{\algorithmicrequire}{ \textbf{Input:}} 
\renewcommand{\algorithmicensure}{ \textbf{Output:}} 
\newcommand{\cmark}{\ding{51}}%
\newcommand{\xmark}{\ding{55}}%
\newcommand{\ie}{\textit{i}.\textit{e}.}
\newcommand{\vs}{\textit{v}.\textit{s}.}
\newcommand{\etc}{\textit{etc}}
\renewcommand{\emph}[1]{\textit{#1}}

\newcommand{\chiccvadd}[1]{\textcolor{black}{#1}}
\newcommand{\del}[1]{\textcolor{red}{}}
\newcommand{\newadd}[1]{\textcolor{black}{#1}}
\newcommand{\chbmvcadd}[1]{\textcolor{black}{#1}}
\maketitle

\begin{abstract}
The practical value of existing supervised sketch-based image retrieval (SBIR) algorithms is largely limited by the requirement for intensive data collection and labeling. In this paper, we present the first attempt at unsupervised SBIR to remove the labeling cost (both category annotations and sketch-photo pairings) that is conventionally needed for training. Existing single-domain unsupervised representation learning methods perform poorly in this application, due to the unique cross-domain (sketch and photo) nature of the problem. We therefore introduce a novel framework that simultaneously performs sketch-photo domain alignment and semantic-aware representation learning. Technically this is underpinned by introducing joint distribution optimal transport (JDOT) to align data from different domains, which we extend with trainable cluster prototypes and feature memory banks to further improve scalability and efficacy. Extensive experiments show that our framework achieves excellent performance in the new unsupervised setting, and performs comparably to existing zero-shot SBIR methods. 
\end{abstract}

\section{Introduction}
Sketches efficiently convey the shape, pose and fine-grained details of objects, and thus are particularly valuable in 
serving as queries to conduct retrieval of photos, \ie, sketch-based image retrieval \cite{sangkloy2016sketchy,yu2016sketch} (SBIR). SBIR has been increasingly well studied, leading to continual improvements in retrieval performance \cite{song2017deep,bhunia2020sketch}. However state-of-the-art methods generally bridge the sketch-photo domain gap through supervised learning using sketch-photo pairs and class annotation \cite{yu2016sketch,sangkloy2016sketchy}. This supervised learning paradigm imposes a severe bottleneck on the feasibility of SBIR in practice. \chbmvcadd{One main research direction on reducing annotation cost thus far has been zero-shot (category generalized) SBIR \cite{dey2019doodle,pang2019generalising,wang2021norm}, where labeled data is no longer necessitated for unseen categories, yet the problem still requires availability of all category labels and specific pairing annotations for the \emph{seen} categories. Furthermore, \cite{radenovic2018deep} turns images into edge maps to directly mitigate the domain gap, but automatically generated pairs from 3D models are still prerequisites to facilitate effective SBIR.}

\begin{figure}[t]
\centering
\includegraphics[width=0.45\columnwidth]{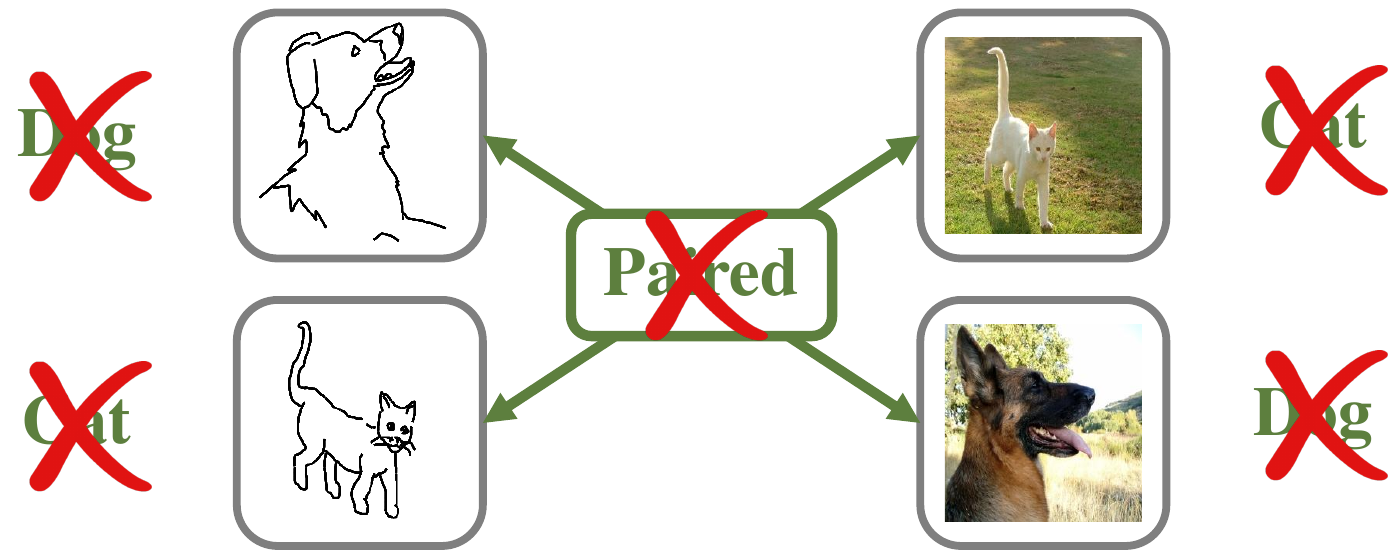}
\caption{Illustration of unsupervised SBIR where no class label or pairing information is available during training.}
\label{illustration}
\end{figure}

\begin{figure}[h!]
\centering
\includegraphics[width=1\columnwidth]{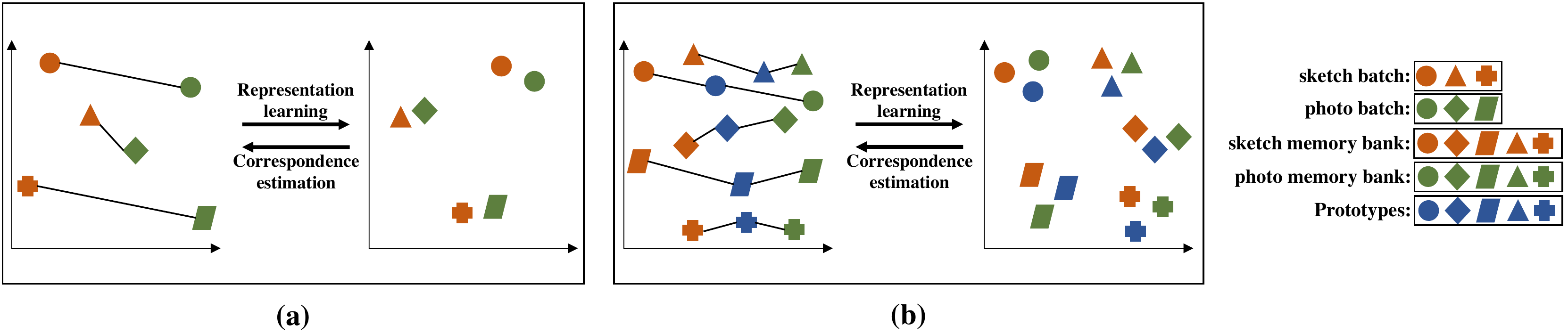}
\caption{\chiccvadd{Comparison between batch-wise DeepJDOT \cite{damodaran2018deepjdot} and our PM-JDOT. Shapes represent samples of different classes. There are five classes in total for demonstration purposes. (a) In batch-wise DeepJDOT, a single batch only contains samples from a subset of classes, so correspondence is necessarily inaccurate (mismatched shapes/categories linked) and poor alignment is learned. (b) In our PM-JDOT, (i) correspondence is mediated by learned prototypes (blue) for all classes, which enables accurate and efficient computation compactly summarizing the whole distribution; (ii) use of a memory bank allows larger sample size with more unique categories than a single batch, increasing the chance that accurate correspondence can be discovered. Note that hard pairwise correspondence is shown for ease of visualization, but actual OT correspondence computation is soft many-many.}}
\label{fig:pm_jdot_illus1}
\end{figure}
In this paper we go to the extreme in addressing the annotation bottleneck, and study for the first time the problem of \chiccvadd{\emph{unsupervised} category-level SBIR, where we work under the stringent assumption of (i) no sketch-photo pairing, and (ii) no category annotations (as illustrated in Figure~\ref{illustration}}) to retrieve photos of the same category as input sketch. We are largely inspired by the recent rapid progress in unsupervised representation learning for photo recognition \cite{chen2020simple,caron2020unsupervised}. However these methods are unsuited to SBIR for the key reason that they are designed for \chiccvadd{single-domain} (photo) representation learning, while SBIR involves \chiccvadd{\emph{cross-domain}} data with a mixture of realistic photos and abstract/iconic sketches. \chiccvadd{Successful category-level SBIR has thus far relied on sketch-photo pairings and category annotations to drive explicit sketch-photo domain alignment and class-discriminative feature learning prior to retrieval \cite{sangkloy2016sketchy,yu2016sketch,pang2019generalising,pandey2020stacked,dutta2019semantically}. The key question for us is  how such alignment and representation learning can be induced just by working with raw \textit{unpaired and unannotated} photos and sketches.} 

At a high-level our solution is based on alternating optimization between: (i) computing a soft (many-to-many) correspondence between sketch and photo domains
; and (ii) learning a representation that aligns sketch and photo features under the soft correspondence, and is also semantically meaningful. Our framework is significantly more performant and resistant to local minima compared to hard noisy pairing methods in other applications \cite{fu2019self, zhang2019self} due to soft correspondence prediction; and a multi-task representation learning objective that synergistically combines cross-domain alignment and in-domain self-supervision for domain-agnostic and semantic-aware feature learning.

In more detail, we first introduce a novel cluster Prototype and feature Memory bank-enhanced Joint Distribution Optimal Transport (PM-JDOT) algorithm for accurate soft cross-domain correspondence estimation. The vanilla JDOT learns to predict cross-domain correspondence using distribution-level information by OT \cite{villani2008optimal}. However, the application of vanilla JDOT in CNNs  \cite{damodaran2018deepjdot} suffers from an inability to simultaneously provide efficiency and accuracy: OT correspondence is either inaccurate if computed at minibatch level (\eg, a given sketch+photo minibatch likely contains a disjoint set of categories, and thus cannot be correctly aligned, as illustrated in Figure~\ref{fig:pm_jdot_illus1}(a)); or intractable if computed at dataset level due to $O(N^2)$ cost. We elegantly solve both of these problems by computing OT between cluster prototypes and instances in feature memory bank, which provide a sparse representation of the full dataset; and extending JDOT with features in memory bank to aggregate information across batches (Figure~\ref{fig:pm_jdot_illus1}(b)). To capture domain-invariant yet class-discriminative features for effective SBIR, we devise an alignment loss to minimize the cross-domain feature discrepancy according to the predicted soft sketch-photo correspondence, and employ a self-supervised loss that helps encode discriminative semantic features by preserving the consistency in cluster assignments between different variants of the same input.

Our main contributions are summarized as follows: (i) We provide the first study of unsupervised SBIR.  (ii) We propose a novel unsupervised learning algorithm for multi-domain data that jointly performs cross-domain alignment and semantic-aware feature encoding. (iii) The cluster prototypes and feature memory banks introduced by our PM-JDOT algorithm alleviates the limits of existing JDOT, enabling effective yet tractable distribution alignment. (iv) Extensive experiments on Sketchy-Extended and TUBerlin-Extended datasets illustrate the promise of our framework in both unsupervised and zero-shot SBIR settings.
\section{Related Work}
\keypoint{Sketch-based image retrieval}
SBIR methods can be classified into two groups according to granularity: Category-level SBIR aims to rank photos so that those with the same semantic class as the input sketch appear first. Fine-grained SBIR targets on retrieving the specific photo corresponding to the query instance. Traditional supervised SBIR algorithms learn class-discriminative feature using classification loss \cite{sangkloy2016sketchy} and remedy the domain gap with sketch-photo paired data \cite{yu2016sketch, song2017deep,bhunia2022adaptive,bhunia2022sketching}. On account of the data shortage that results from\cut{ time-consuming and} labour-intensive sketch-photo paired dataset collection and annotation, zero-shot SBIR intends to test on novel categories that are unseen during training. Representative approaches use adversarial training strategy \cite{dutta2019semantically, pandey2020stacked} or triplet ranking loss \cite{yelamarthi2018zero,sain2022sketch3t} to learn a common feature space for both domains. Additional side information like word embeddings \cite{dey2019doodle} may also be exploited  to preserve semantic information. Nevertheless, annotated training data is still necessary in existing zero-shot SBIR approaches to perform effective training, and the required cross-category generalization is still an active research question \cite{pang2019generalising}. We are therefore motivated to study unsupervised category-level SBIR that does not rely on sketch-photo annotations. 

\keypoint{Unsupervised deep learning}
Unsupervised deep learning methods have recently made strong progress in representation learning that ultimately diminishes the demand for data annotation. Most  contemporary unsupervised learning methods can be classified into four categories according to the learning objective: (i) Deep clustering approaches are designed to model the feature space via data grouping where the pseudo class label can be assigned with the help of clustering algorithm \cite{caron2018deep, gao2020deep}. (ii) Instance discrimination \cite{wu2018unsupervised} treats every single sample as a unique class, which can be beneficial to capture discriminative features of individual instance. (iii) Self-supervised learning algorithms learn through solving different pretext tasks including image colorization \cite{zhang2016colorful}, image super-resolution \cite{ledig2017photo}, image in-painting \cite{pathak2016context}, solving jigsaw puzzle \cite{noroozi2016unsupervised}, rotation prediction \cite{gidaris2018unsupervised}. (iv) Contrastive learning aims to maximize agreement between different augmentations of the same input in feature space \cite{chen2020simple} or label space \cite{caron2020unsupervised}. However, these methods are designed for single domain  representation learning, and perform poorly if applied directly to multi-domain data. A few self-supervised methods have been defined for multi-domain data \cite{tian2019contrastiveCMC}, but these normally assume that cross-domain pairing is the `free' pre-text task label, which is exactly the annotation we want to avoid. In contrast, our model performs unsupervised learning in each domain, while simultaneously aligning the domains through JDOT.  

\keypoint{Joint distribution optimal transport}
Optimal transport (OT) \cite{villani2008optimal} is a mathematical theory that enables distance measurement between distributions by way of searching for the optimal transportation plan to match samples from both distributions. OT has been applied in domain adaptation \cite{courty2014domain, perrot2016mapping, yan2018semi} to learn a transportation plan between source and target domains, followed by training a classifier for target domain with transported source domain data and the corresponding category annotation. To avoid this two-step process (feature transformation and classification model training), JDOT \cite{courty2017joint} aligns the feature-label joint distribution and projects input samples from both domains onto a common feature space where a
classifier can be shared. DeepJDOT \cite{damodaran2018deepjdot}  extends JDOT to deep learning and facilitates  training on large scale datasets by introducing a stochastic approximation via batch-wise OT. However, we observe that data in a single batch is not informative enough to represent the whole data distribution, which limits the efficacy of OT in DeepJDOT. To this end, we introduce PM-JDOT which employs prototypes and feature memory banks to enhance representation of each distribution for optimizing OT-based cross-domain alignment. 

\section{Methodology}
In category-level SBIR, the goal is to train an effective CNN $f_\theta:I\rightarrow \mathbf{x}$ to project input imagery $I$ from both sketch and photo domains into a shared embedding space, where features $\mathbf{x}$ facilitate cross-domain instance similarity measurement. \chiccvadd{Given a query sketch $I^s$, a ranked list of photos will be generated according to their feature space distance to the query with the aim of ranking photos of the same category on top of the list.} In the proposed unsupervised setting, we only have access to a set of training sketches $\mathcal{I}^s = {\left\{I_i^s\right\}}_{i=1}^M$ and photos $\mathcal{I}^p= {\left\{I_j^p\right\}}_{j=1}^N$ \chiccvadd{that contain the same categories, but without category or sketch-photo pairing annotations  -- thus raising the challenge of how to learn a representation suitable for retrieval.}

\begin{figure*}
\centering
\includegraphics[width=1\columnwidth]{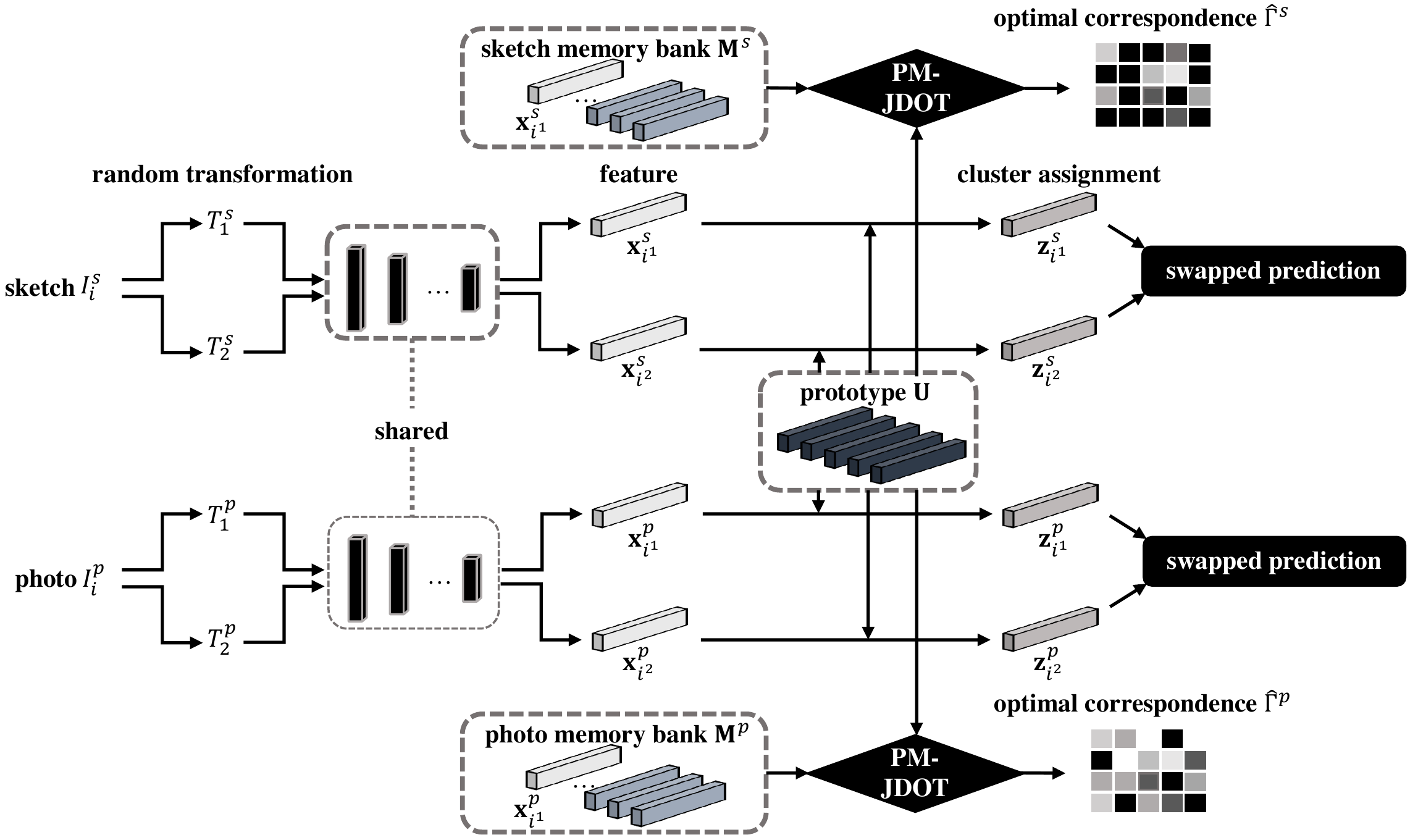}
\caption{Schematic of our proposed framework. \cut{Sketches and photos are projected into a common feature space by unsupervised representation learning and distribution alignment.  which transforms sketch and photo into a common feature space for image retrieval. Feature extractor and prototype are shared by both sketch and photo.}}
\label{overview}
\end{figure*}

To solve this problem, our method integrates two objectives: (i) cross-domain correspondence estimation with PM-JDOT, which employs \chiccvadd{the aggregated data in} trainable cluster prototypes and feature memory banks in support of accurate and scalable discrepancy measurement. 
and (ii) unsupervised feature representation learning that encodes domain-agnostic and semantic-discriminative features from visual input.
Figures~\ref{overview} briefly summarizes our unsupervised SBIR framework.

\subsection{Cross-domain correspondence estimation}
\label{section:corres-est}
\keypoint{From JDOT to PM-JDOT} 
Given only unpaired and unlabeled photos and sketches, we introduce a machinery to estimate the soft sketch-photo correspondence in an unsupervised way to support the cross-domain alignment. We introduce joint distribution optimal transport (JDOT) to match samples from the sketch and photo domains. Crucially, we extend it to improve both alignment accuracy and efficiency by redefining the problem in terms of OT between a set of learnable prototypes and feature memory banks -- PM-JDOT.
Conventional JDOT is able to align all features $\left\{\mathbf{x}_i^s\right\}_{i=1}^M$ from sketch domain and $\left\{\mathbf{x}_j^p\right\}_{j=1}^N$ from photo domain via computing the optimal transport plan between them. To make this quadratic computation scale to \chiccvadd{neural network training}, JDOT is applied \chiccvadd{between two randomly selected batches} \cite{damodaran2018deepjdot}. However, an individual sketch/photo batch is a \chiccvadd{weak representation for the overall data distribution of one domain, leading to poor correspondence 
as illustrated in Figure~\ref{fig:pm_jdot_illus1}}.  

Thus we first exploit $K$ trainable cluster prototypes $\mathbf{U} = [\mathbf{u}_1, \mathbf{u}_2, ... , \mathbf{u}_K]$ as a stronger proxy to learn a better alignment. Specifically, instead of matching sketch and photo \chiccvadd{batches in isolation}, we estimate the correspondence between sketch/photo batches and the prototypes which compactly represent the whole dataset 
with a small number of elements. 
\del{Furthermore, batch-wise JDOT represents the full distribution in each domain in the impoverished form of a small batch of samples. To alleviate this limitation, }\chiccvadd{To further alleviate the limitation caused by the impoverished domain representation, \ie, a small batch of samples, }we introduce feature memory banks of size $E$ for sketch $\mathbf{M}^s = [\mathbf{x}^s_1, \mathbf{x}^s_2, ..., \mathbf{x}^s_E]$ and photo $\mathbf{M}^p = [\mathbf{x}^p_1, \mathbf{x}^p_2, ..., \mathbf{x}^p_E]$ \del{to replace single batches}
\chiccvadd{as richer domain representations which augment current batch with samples in previous batches.} The update strategy of memory bank is FIFO, \ie, removing the oldest batch and putting the current batch on the top of the container. 

\keypoint{Correspondence search} \chiccvadd{In PM-JDOT}, the correspondence $\Gamma$ is found from the set of transportation plans $\Pi$ between prototypes and one feature memory bank by minimizing\del{ the displacement cost}:
\begin{equation}
\label{couple_select-eq}
\begin{split}
&\min_{\Gamma \in \Pi} \conghuiadded{\sum_{i=1}^{K} \sum_{j=1}^{E}} \ \Gamma_{ij}\mathbf{C}(i, j) - \lambda H(\Gamma) , \quad \text{where}\\
&\Pi = \left\{\Gamma \in \mathbb{R}_{+}^{K \times E} | \Gamma\mathbf{1}_{E} = \frac{1}{K}\mathbf{1}_{K}, \Gamma^{\top}\mathbf{1}_{K} = \frac{1}{E}\mathbf{1}_{E} \right\}
\end{split}
\end{equation}
Here, $H(\cdot)$ is an entropy regularization term weighted by $\lambda$. $K$ and $E$ are the number of prototypes and the size of feature memory bank respectively. The constraint for transportation plans $\Pi$ ensures each prototype can be selected $\frac{E}{K}$ times on average \cite{caron2020unsupervised}. $\mathbf{C} \in \mathbb{R}^{K \times E}$ is the matrix of cross-domain pairwise costs. Specifically, the cost $\mathbf{C}(i, j)$ of aligning $i^{th}$ prototype and $j^{th}$ 
sample in memory bank is calculated by:
\begin{equation}
\label{jdot_cost_item-eq}
\begin{split}
\mathbf{C}(i, j) &= \alpha d_{f}(\mathbf{u}_i, \mathbf{x}_j) + \beta d_{l}(\mathbf{v}_i, \mathbf{y}_j), \quad \text{where} \\ 
\mathbf{y}_{j}^{(k)}\cut{(\tau_1)} &= \frac{\exp(\mathbf{x}^{\top}_{j}\mathbf{u}_k/{\conghuiadded{\tau}})}{\sum\limits_{m=1}^{K}\exp(\mathbf{x}^{\top}_{j}\mathbf{u}_m/{\conghuiadded{\tau}})}
\end{split}
\end{equation}
\chiccvadd{Here, cosine distance $d_{f}$ is used to measure the feature-wise similarity between prototype vector $\mathbf{u}_i$ and feature $\mathbf{x}_j$ extracted with $f_\theta$. $d_{l}$ is applied label-wise to evaluate the difference between one-hot label $\mathbf{v}_i$ for $i^{th}$ prototype and cluster probability $\mathbf{y}_j$ for $j^{th}$ image in the memory bank. $\mathbf{v}_i$ is generated automatically according to the index $i$, \eg, $\mathbf{v}_1 = [0, 1, 0, 0, ... , 0]$.} $\alpha$ and $\beta$ are scalar hyperparameters that control the contributions of feature and label distance measurements. PM-JDOT is executed twice for prototype-sketch and prototype-photo correspondence, producing optimal correspondence $\hat{\Gamma}^s$ and $\hat{\Gamma}^p$ respectively. Feature extractor $f_\theta$ and prototypes $\mathbf{U}$ are fixed in this process.

\subsection{Unsupervised representation learning}
\label{section:unsup-learn}
The algorithm so far in Section~\ref{section:corres-est} learns the correspondence between prototypes and samples in both domains, but the feature extractor is not optimized. Thus, in this section, we further illustrate the second part of our alternating optimization: unsupervised representation learning to train $f_\theta$ to extract features that are domain-invariant (aligned across domains), yet sensitive to semantic category. 




\keypoint{Cross-domain alignment} In order to align features from sketch and photo domains, \chiccvadd{we leverage the first $A$ columns in  $\hat{\Gamma}^s$ and $\hat{\Gamma}^p$, which contain the mapping between trainable prototypes and current sketch/photo batch of size $A$.}\del{we need to extract the optimal transportation plans related to current batch from $\hat{\Gamma}^s$ and $\hat{\Gamma}^p$. Specifically, from $\hat{\Gamma}^s \in \mathbb{R}_{+}^{K \times E}$ and $\hat{\Gamma}^p \in \mathbb{R}_{+}^{K \times E}$, the first $A$ columns corresponding to the transportation plan between prototypes and current batch of size $A$ are selected.} Then the feature extractor $f_\theta$ and trainable prototypes $\mathbf{U}$ are updated by minimizing feature and label discrepancy between corresponding prototypes and samples in the batch \chiccvadd{according to the optimal correspondences}:
\begin{equation}
\label{jdot_f_cost-eq}
\begin{aligned}
L_{a} = &L^s_{a} + L^p_{a} \\[-1pt]
= &(\sum_{i=1}^{K}  \sum_{j=1}^{A} \hat{\Gamma}^s_{ij}(\alpha d_{f}(\mathbf{u}_i, \mathbf{x}_j^s) + \beta d_{\conghuiadded{l}}(\mathbf{v}_i, \mathbf{y}_j^{s})))\\[-1pt]
&+(\sum_{i=1}^{K}  \sum_{j=1}^{A} \hat{\Gamma}^p_{ij}(\alpha d_{f}(\mathbf{u}_i, \mathbf{x}_j^p) + \beta d_{\conghuiadded{l}}(\mathbf{v}_i, \mathbf{y}_j^{p})))
\end{aligned}
\end{equation}

\keypoint{Semantic-aware feature learning} To learn a semantically meaningful representation  (\ie, ensure samples from same category are similar in feature space) from unannotated pixel-level input images, 
inspired by SwAV \cite{caron2020unsupervised}, we train feature extractor $f_\theta$ by 
contrasting the cluster assignments \chiccvadd{for different variants of the same image}. 
The training objective is to minimize the semantic representation loss\del{$L_{se}$ with respect to $\theta$ and trainable prototypes $\mathbf{U}$}:
\begin{equation}
\label{swav-eq}
\begin{aligned}
L_{se} &= L^s_{se} + L^p_{se} \\
&= (\ell(\mathbf{y}_{i^1}^{s}, \mathbf{z}_{i^2}^{s}) + \ell(\mathbf{y}_{i^2}^{s}, \mathbf{z}_{i^1}^{s})) + (\ell(\mathbf{y}_{i^1}^{p}, \mathbf{z}_{i^2}^{p}) + \ell(\mathbf{y}_{i^2}^{p}, \mathbf{z}_{i^1}^{p})) \\
\end{aligned}
\end{equation}
where $\ell$ is the cross-entropy loss. Taking sketch domain for illustration, $\mathbf{y}_{i^t}^{s}$ and $\mathbf{z}_{i^t}^{s}$ are predicted cluster probability and cluster assignment of $I_{i^t}^{s}$ respectively. $\{\mathbf{y}_{i^1}^{s}, \mathbf{z}_{i^1}^{s}\}$ and $\{\mathbf{y}_{i^2}^{s}, \mathbf{z}_{i^2}^{s}\}$ correspond to two transformed variants $I_{i^1}^{s} = T_1(I_i^{s})$ and $I_{i^2}^{s}=T_2(I_i^{s})$ of the same original sketch $I_i^{s}$, where $T_1$ and $T_2$ are randomly sampled from the set $\mathcal{T}$ of image transformations including rescaling, flipping, \etc. Through \emph{swapped prediction}, \ie, pairing $\mathbf{y}_{i^1}^{s}$ with $\mathbf{z}_{i^2}^{s}$ and $\mathbf{y}_{i^2}^{s}$ with $\mathbf{z}_{i^1}^{s}$ in cross-entropy loss $\ell$, the network learns to predict consistent cluster probabilities for different augmentations of identical image, which assists semantically-aware feature learning. 
$\mathbf{y}_{i^t}^{s}$ can be measured in the same way as Equation~\ref{jdot_cost_item-eq}. 
And we compute cluster assignment $\mathbf{z}_{i^t}^{s}$ online at each iteration as follow \cite{caron2020unsupervised}:
\begin{equation}
\label{z-eq}
\begin{split}
&\max_{\mathbf{Z} \in \mathcal{Z}} \text{Tr}(\mathbf{Z}^{\top}\mathbf{U}^{\top}\mathbf{Q}) + \epsilon H(\mathbf{Z}) , \quad \text{where}\\
&\mathcal{Z} = \left\{\mathbf{Z} \in \mathbb{R}_{+}^{K \times B} | \mathbf{Z}\mathbf{1}_{B} = \frac{1}{K}\mathbf{1}_{K}, \mathbf{Z}^{\top}\mathbf{1}_{K} =  \frac{1}{B}\mathbf{1}_{B} \right\}
\end{split}
\end{equation}
\chiccvadd{Where $\mathbf{Q}$ is a feature queue of size $B$ which is initialized with image features and updated continuously in a FIFO manner during training. If the training batch size is $A$, the current batch features define the top $A$ elements in $\mathbf{Q}$. $\mathbf{Z}$ are cluster assignments corresponding to the $B$ samples in $\mathbf{Q}$. $\mathbf{U}$ represents cluster prototypes. 
$H(\cdot)$ is an entropy penalty with weight $\epsilon$. 
Only the cluster assignments for current batch, \ie, top $A$ elements in $\mathbf{Z}$, are used for $L_{se}$.}

\keypoint{Summary} The overall learning objective is to train an effective feature extractor $f_\theta$ 
without class or instance-paired annotation. We achieve this by minimizing alignment loss $L_{a}$ and the semantic representation loss $L_{se}$ as:
\begin{equation}
\label{total_cost-eq}
\begin{aligned}
\underset{\theta, \mathbf{U}}{\operatorname{argmin}} \ \nu L_{a} + \mu L_{se}
\end{aligned}
\end{equation}
where $\nu$ and $\mu$ are respective loss weights. Algorithm~1 in Supplementary material summarizes the training algorithm followed in this work. 






\section{Experiments}
\subsection{Datasets and Settings}
\keypoint{Datasets} We evaluate our algorithm  on two datasets: (i) Sketchy-Extended \cite{liu2017deep} contains 75,471 free-hand sketches and 12,500 photos spanning over 125 categories provided by \cite{sangkloy2016sketchy} and another 60,502 photos collected in \cite{liu2017deep} from ImageNet \cite{deng2009imagenet}. (ii) TUBerlin-Extended \cite{zhang2016sketchnet} offers 20,000 sketches \cite{eitz2012humans} evenly distributed on 250 classes and photos of same categories collected using Google image search. 

\keypoint{Implementation details} We use ResNet-50 \cite{he2016deep} as feature extractor $f_\theta$, followed by an additional L2 normalization layer to transform visual input into 128-d feature embeddings. $f_\theta$ is first initialized with parameters pre-trained with photos in ImageNet dataset \cite{deng2009imagenet} by applying SwAV \cite{caron2020unsupervised}. \newadd{As SwAV \cite{caron2020unsupervised} is an unsupervised learning framework, it is guaranteed that \textit{no labeled data is used in the pre-training stage}.} \chiccvadd{All training photo features extracted with the pre-trained $f_\theta$ are grouped into $K$ clusters using K-means. The $K$ cluster centroids are then employed to initialize prototypes $\mathbf{U}$.} The number of prototypes $K$ is set to the actual number of training categories, \ie, 125 for Sketchy-Extended and 250 for TUBerlin-Extended in unsupervised SBIR. The sum of elements related to current batch in $\hat{\Gamma}^s$ and $\hat{\Gamma}^p$ are normalized to 1\cut{ensure the sum of the whole matrix is 1} in Equation~\ref{jdot_f_cost-eq} for all experiments\cut{ to facilitate fair comparisons}. Both the feature extractor and prototypes are trained with learning rate initialized with 0.01 and divided by 2 after each 10 epochs. We use SGD optimizer and set momentum factor and weight decay value to 0.9 and 1e-4 respectively. Weights for $L_{a}$ and $L_{se}$ are 1 and 10. And temperature hyperparameter $\tau$ is set to 0.1. Our framework is implemented with Pytorch \cite{paszke2019pytorch} and optimal transportation plans are computed by the POT toolbox \cite{flamary2017pot}.

\keypoint{Evaluation metrics} Cross-domain retrieval is performed by computing cosine distance between sketch and photo feature vectors and generating a ranked list of gallery photos. We evaluate the retrieval performance by calculating the precision and mean average precision among top 200 retrieved photos denoted by Prec@200 and mAP@200 as well as the mean average precision over the whole dataset (mAP). Photos belonging to the same category as the query sketch are considered as correct retrievals.
\begin{table*}[t]
\begin{center}
\caption{Unsupervised SBIR results on Sketchy-Extended and TUBerlin-Extended dataset}
\scalebox{0.68}{
\begin{tabular}{r|c|c|c|c|c|c}
\hline
\multirow{2}{*}{Method}
& \multicolumn{3}{c|}{Sketchy-Extended dataset} & \multicolumn{3}{c}{TUBerlin-Extended dataset}\\
\cline{2-7}
& Prec@200$(\%)$ & mAP@200 $(\%)$ & mAP $(\%)$ & Prec@200$(\%)$ & mAP@200 $(\%)$ & mAP $(\%)$ \\
\hline
RotNet \cite{gidaris2018unsupervised} & 2.26 & 4.89 &1.54 & 1.53 & 3.61 & 0.77 \\
ID \cite{wu2018unsupervised} & 3.41 &5.26 & 2.45 & 2.66 & 5.35 & 1.35 \\
CDS \cite{kim2020cross} & 2.37 & 3.58 & 1.88 & 2.64 & 4.69 & 1.63  \\
GAN \cite{goodfellow2014generative} & 2.45 & 4.66 & 1.43  & 1.56 & 3.45 & 0.69 \\
SwAV \cite{caron2020unsupervised} & 10.87 & 12.51 & 10.15 & 3.36 & 5.81 & 2.89 \\
DSM \cite{radenovic2018deep} & 10.07 & 17.92 & 4.28 & 7.05 & 13.00 & 2.61 \\
SwAV \cite{caron2020unsupervised} + CycleGAN \cite{CycleGAN2017} & 4.15 &  5.39 & 4.28 & 2.67 & 3.50 & 2.06 \\
SwAV \cite{caron2020unsupervised} + GAN \cite{goodfellow2014generative}& 22.96 & 25.48 & 18.82 &10.92 & 13.46 & 8.43 \\
Ours & \textbf{33.64} & \textbf{36.31} & \textbf{28.17} & \textbf{14.78} & \textbf{18.66} & \textbf{9.93} \\
\hline
\end{tabular}
}
\label{compare-retrieval}
\end{center}
\vspace{-1.5em}
\end{table*}

\subsection{Results}
\subsubsection{Unsupervised SBIR}
\keypoint{Settings}
50 and 10 sketches for each class are randomly selected as query sets for Sketchy-Extended and TUBerlin-Extended dataset respectively for testing. The remaining sketches and photos are used during the training process by following the same setting in \cite{liu2017deep}. No category labels or sketch-photo pairings are available during training. Each mini-batch contains 128 96 $\times$ 96 pixel sketches and photos. 

\begin{figure}[t]
\centering
\includegraphics[width=1\linewidth]{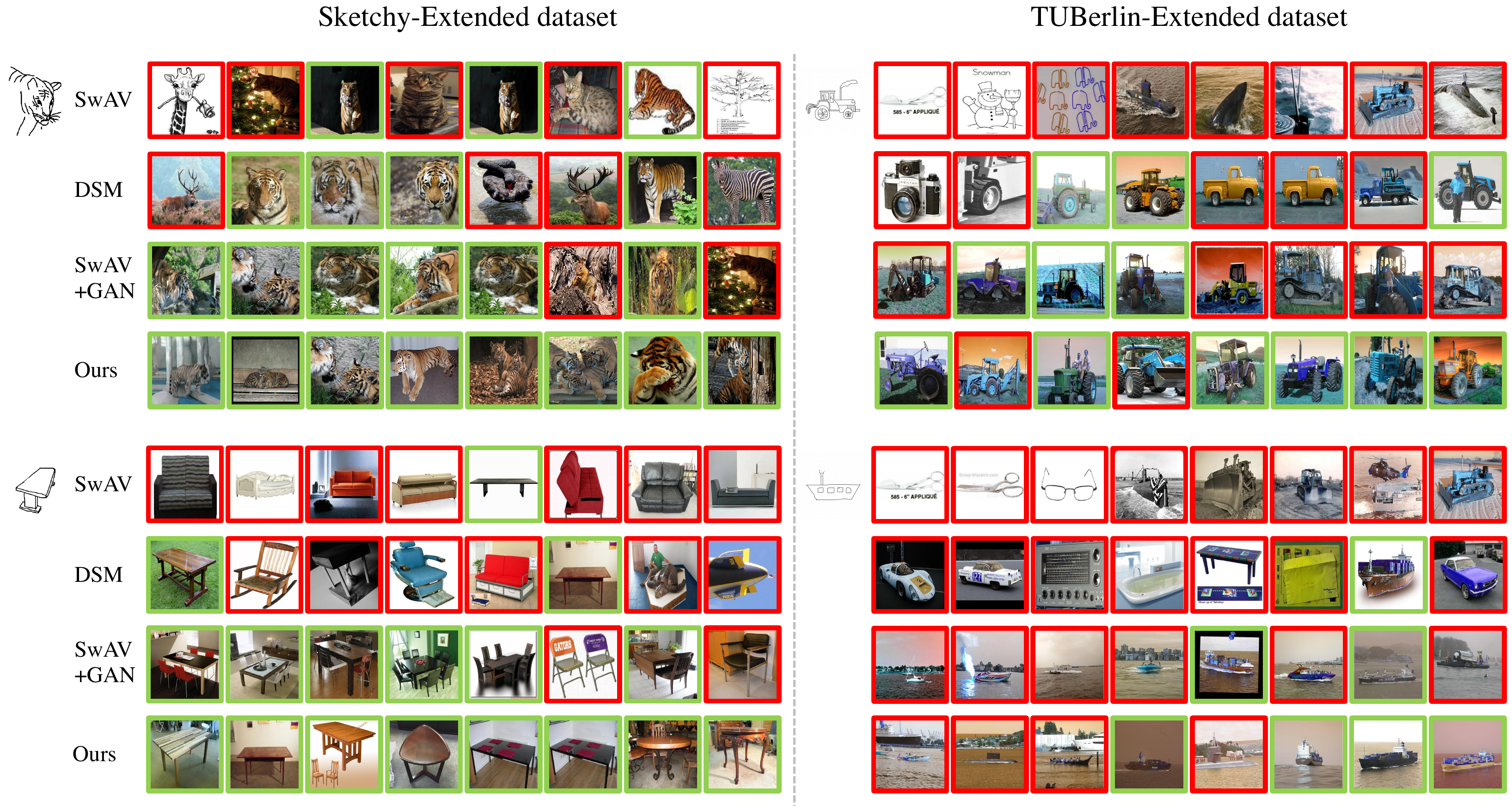}
\caption{Top8 retrieval results for unsupervised SBIR. Row 1\&5: Retrieval results of SwAV \cite{caron2020unsupervised}; Row 2\&6: Retrieval results of DSM \cite{radenovic2018deep}; Row 3\&7: Retrieval results of SwAV \cite{caron2020unsupervised} + GAN \cite{goodfellow2014generative}; Row 4\&8: Retrieval results of our framework.}
\label{qualitative-result}
\vspace{0.3em}
\end{figure}

\begin{figure}[t]
\centering
\includegraphics[width=1\columnwidth]{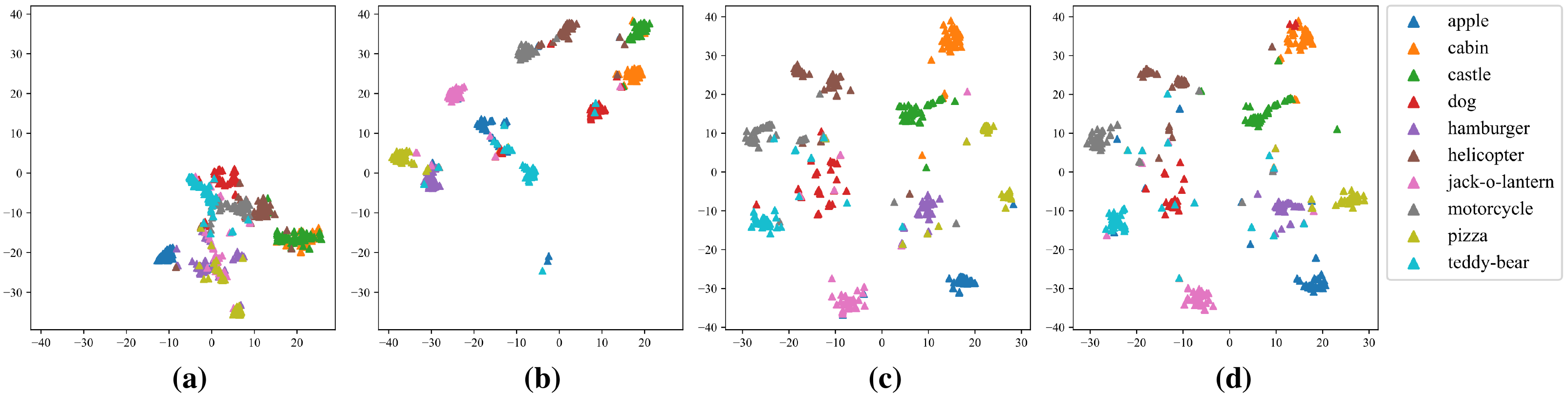}
\caption{\textit{t-SNE} visualization of 10 categories from Sketchy-Extended dataset. (a): Sketch feature visualization of SwAV \cite{caron2020unsupervised}; (b): Photo feature visualization of SwAV \cite{caron2020unsupervised}; (c): Sketch feature visualization of our method; (d) Photo feature visualization for our method.}
\label{feat vis-result}
\end{figure}
\keypoint{Results}
Quantitative retrieval results on Sketchy-Extended and TUBerlin-Extended are shown in Table \ref{compare-retrieval}. From the results, we make the following observations: (i) Unsupervised feature representation learning algorithms \cite{gidaris2018unsupervised, wu2018unsupervised, caron2020unsupervised} originally designed for single-domain perform poorly when directly applied to cross-domain task like SBIR. SwAV \cite{caron2020unsupervised} is the best among these three methods. (ii) CDS \cite{kim2020cross} cannot cope with the large domain gap between sketch and photo and results in unsatisfactory performance. (iii) From the comparison between GAN \cite{goodfellow2014generative} and SwAV+GAN, we can see that additional guidance targeting on preserving semantic-discriminative feature is essential in category-level SBIR. (iv) CycleGAN fails to generate high-quality color images from sketch in large-scaled multi-class image translation. In contrast, it degrades the semantic information in the original sketch and leads to worse retrieval results compared with SwAV only. (v) Our proposed framework achieves the best retrieval accuracy compared with all these baseline methods trained without external labeled data. Quantitative retrieval results and feature visualizations can be found in Figure \ref{qualitative-result} and Figure \ref{feat vis-result}.


\begin{table*}[t]
\begin{center}
\caption{Zero-shot SBIR results on Sketchy-Extended and TUBerlin-Extended dataset. $(*)^a$ represents for retrieval results on 25 test categories following the setting proposed in \cite{saavedra2015sketch}. All methods except ours use instance-wise annotation in the train set.}
\vspace{-0.5em}
\scalebox{0.68}{
\begin{tabular}{r|c|c|c|c|c|c|c}
\hline
\multirow{2}{*}{Method} & \multirow{2}{*}{Supervision}
& \multicolumn{3}{c|}{Sketchy-Extended dataset} & \multicolumn{3}{c}{TUBerlin-Extended dataset}\\
\cline{3-8}
&& Prec@200$(\%)$ & mAP@200 $(\%)$ & mAP $(\%)$ & Prec@200$(\%)$ & mAP@200 $(\%)$ & mAP $(\%)$ \\
\hline
ZSIH \cite{shen2018zero} &\cmark & -& - & 25.90$^a$ & - & - & 23.40  \\

CVAE \cite{yelamarthi2018zero}&\cmark &33.30 & 22.50 &19.59 & 0.30 & 0.90 & 0.50 \\

SAN \cite{pandey2020stacked}&\cmark & 32.20 & 23.60 & - & 21.80 & 14.10 & - \\

SEM-PCYC \cite{dutta2019semantically}&\cmark & - & - & 34.90$^a$ & -  & - & \textbf{29.70} \\

Doodle \cite{dey2019doodle}&\cmark & 37.04 & \textbf{46.06} & \textbf{36.91}  & 12.08 & 15.68 & 10.94 \\

Ours&\xmark & \textbf{38.44}  & 44.09 & 34.68 & \textbf{28.36} & \textbf{31.53} & 22.91 \\
\hline
\end{tabular}
}
\label{zs-sbir res}
\end{center}
\end{table*}

\subsubsection{Zero-shot SBIR}
\keypoint{Settings} We use the same data split as \cite{dey2019doodle}: 104 and 21 categories are selected for training and testing respectively for Sketchy-Extended dataset. 30 classes are randomly chosen from TUBerlin-Extended dataset for testing and the rest are used for training. Following the default  setting in \cite{dey2019doodle}, we set each mini-batch to 20 224 $\times$ 224 sketches and photos.


\keypoint{Results} Retrieval performance in Table \ref{zs-sbir res} shows that even without involving human pairwise or category-level annotations during training, our framework still performs comparably with existing zero-shot SBIR algorithms that use such annotations during training. Our aligned semantically rich and domain-invariant representation learned on unlabeled training data can generalize directly to unseen classes not used for training. 
\begin{table*}[t]
\begin{center}
\caption{Ablation study on our model components. Unsupervised SBIR  on Sketchy-Extended and TUBerlin-Extended dataset.}
\vspace{-0.5em}
\scalebox{0.67}{
\begin{tabular}{c|ccc|c|c|c|c|c|c}
\hline
\multirow{2}{*}{Method}&\multirow{2}{*}{JDOT}&\multirow{2}{*}{Proto.}&\multirow{2}{*}{\shortstack{Mem. \\bank}}
& \multicolumn{3}{c|}{Sketchy-Extended dataset} & \multicolumn{3}{c}{TUBerlin-Extended dataset}\\
\cline{5-10}
&&&& Prec@200$(\%)$ & mAP@200 $(\%)$ & mAP $(\%)$ &Prec@200$(\%)$ & mAP@200 $(\%)$ & mAP $(\%)$ \\
\hline
v1 &\xmark  & \xmark & \xmark & 10.87 & 12.51 & 10.15 &3.36 & 5.81 & 2.89 \\

v2 &\cmark & \xmark & \xmark & 21.07 & 23.19 & 18.53 &7.01 & 9.96 & 5.05 \\

v3 &\cmark & \cmark & \xmark & 25.60 & 28.62 & 20.98 & 9.01 & 12.26 &5.62 \\

v4 &\cmark &\xmark & \cmark & 24.83 & 27.67 & 20.78 & 11.71  & 15.66 & 7.53 \\
v5 &\cmark & \cmark & \cmark  & \textbf{33.64} & \textbf{36.31} & \textbf{28.17} & \textbf{14.78} & \textbf{18.66} & \textbf{9.93} \\
\hline
\end{tabular}
}
\label{abl-model analysis}
\end{center}
\vspace{-1em}
\end{table*}

\subsubsection{Ablation Study}
We analyze the efficacy of different components in our unsupervised SBIR framework in Table~\ref{abl-model analysis}: (i) Compared with vanilla SwAV (v1), JDOT using batch-wise OT (v2) for alignment as in \cite{damodaran2018deepjdot} already benefits cross-domain matching in both datasets; (ii) In v3, the transportation map is measured between prototypes and single batch of instances. The result shows that prototypes offers a better approximation for real data distribution and improves the OT-based alignment; (iii) Making use of additional data for memory bank-wise OT (v4) is also beneficial for feature alignment; and (iv) Our full model (v5), which takes advantages of both prototypes and memory banks, provides best alignment and representation learning strategy. Further analysis can be found in the Supplementary Material.

\section{Conclusion}
This paper presents the first attempt at unsupervised SBIR, which is a more challenging learning problem, but more practically valuable due to addressing the data annotation bottleneck. To facilitate cross-domain feature representation learning with no labeled data, our proposed framework performs cross-domain correspondence estimation and unsupervised representation learning alternatively. Alignment is further achieved accurately and scalably by our PM-JDOT. The results show that our unsupervised framework already provides usable performance on par with contemporary zero-shot SBIR methods, but without requiring any instance-wise category or pairing annotation.

\bibliography{egbib}
\end{document}